\pdfoutput=1

\documentclass[11pt]{article}

\usepackage{coling}

\usepackage{times}
\usepackage{latexsym}
\usepackage[T1]{fontenc}
\usepackage[utf8]{inputenc}
\usepackage{microtype}
\usepackage{inconsolata}
\usepackage{graphicx}
\usepackage[]{mdframed}

\newcommand{\framedtext}[1]{
\par
\noindent\fbox{
    \parbox{\dimexpr\linewidth-2\fboxsep-2\fboxrule}{#1}}
    }

\title{On Crowdsourcing Task Design for Discourse Relation Annotation}

\author{Frances Yung \and
Vera Demberg\\
Saarland University, Saarbr\"ucken, Germany\\
\texttt{\{frances, vera\}@lst.uni-saarland.de}}

\begin{document}
\maketitle
\begin{abstract}
Interpreting implicit discourse relations involves complex reasoning, requiring the integration of semantic cues with background knowledge, as overt connectives like \textit{because} or \textit{then} are absent. These relations often allow multiple interpretations, best represented as distributions. 
In this study, we compare two established methods that crowdsource English implicit discourse relation annotation by connective insertion: a free-choice approach, which allows annotators to select any suitable connective, and a forced-choice approach, which asks them to select among a set of predefined options. Specifically, we re-annotate the whole DiscoGeM 1.0 corpus - initially annotated with the free-choice method - using the forced-choice approach. The free-choice approach allows for flexible and intuitive insertion of various connectives, which are context-dependent.  Comparison among over 130,000 annotations, however, shows that the free-choice strategy produces less diverse annotations, often converging on common labels. Analysis of the results reveals the interplay between task design and the annotators' abilities to interpret and produce discourse relations.
\end{abstract}

\section{Introduction}

Disagreement in linguistic annotation is increasingly seen not as noise but as a valuable signal capturing diverse perspectives in language interpretation \cite{dumitrache2021empirical,uma2021learning,frenda2024perspectivist}. A single gold label, traditionally provided by one or two trained annotators, often fails to capture the full range of interpretations, which may arise from linguistic ambiguity, contextual factors, or annotators’ cultural and experiential backgrounds. Crowdsourcing offers a scalable solution for gathering these alternative interpretations.

To guide untrained crowd workers in reliably annotating abstract linguistic phenomenon, intuitive and carefully designed workflows are essential. Task design has been identified as one of the factors behind annotation disagreement and bias \cite{pavlick2019inherent,jiang2022investigating}, and even can impact annotation quality \cite{shaw2011designing,gadiraju2017clarity,gururangan-etal-2018-annotation}.   
For example, \citet{pyatkin-etal-2023-design} investigate the bias of task design guiding workers in annotating implicit discourse relation (IDR) senses, which often have multiple interpretations. They compared two methods: one based on insertion of discourse connectives (DCs), e.g. \textit{John fell down \textbf{because} he tripped}, and the other on paraphrasing discourse arguments to question-answer (QA) pairs, e.g. \textit{Q: \textbf{Why} (did) John fell down? A: He tripped}. While annotations from both methods were found to align closely, subtle bias in the annotation preference are found in both methods due to limitations of using natural language to annotate specialized linguistic concepts (not all senses can be easily expressed by a connective or by a question). 

Building on this line of work, we explore the potential method bias of two IDR annotation tasks for English based on DC insertion.  These methods differ solely in whether annotators select from predefined options \cite{rohde-etal-2016-filling,yung-etal-2024-discogem} or freely type in their choices \cite{yung-etal-2019-crowdsourcing}. 
The free-choice method was employed to annotate 6,500 English IDRs in the DiscoGeM~1.0 corpus \cite{scholman2022DiscoGeM},whereas the DiscoGeM~2.0 corpus, comprising multi-lingual translations or original texts from DiscoGeM 1.0, was annotated using the forced-choice method.  An initial comparison of the statistics of the two corpora revealed characteristics unique to the English annotations, such as a higher proportion of \textsc{conjunction} relations.
 
Our findings indicate that the free-choice approach achieves higher agreements among annotators, while the forced-choice approach is more effective at capturing a diverse range of alternative interpretations. Further analysis reveals that the free-choice approach favours intuitive and frequent intuitive sense, whereas the provided options in the forced-choice approach serve as prompts for the workers to identify rare, fine-grained senses. Moreover, the method bias interacts with individual differences in discourse processing: workers who could identify a wider range of senses in one approach also tended to label more different senses in the other approach. These results highlight the nuanced impact of task design on annotation outcomes.  

The re-annotated resource is freely downloadable\footnote{\url{https://github.com/merelscholman/DiscoGeM}} alongside the original DiscoGeM 1.0. It provides an interesting dataset for the study of perspectivism and design in annotation as well as a rich collection of rare IDR examples, contributing to the major data bottleneck for current IDR recognition models.

\section{Related work}
Annotation of IDR requires integrating subtle semantic cues with background knowledge and mapping these to abstract labels -- a task that is challenging even for trained annotators \cite{hoek2017evaluating}. Previous attempts to create datasets by crowdsourcing annotations often compromise on label variety or annotation quality \cite{kawahara-etal-2014-rapid,kishimoto-etal-2018-improving}.  

Inspired by the Penn Discourse Treebank's (PDTB) lexicalized approach to annotate IDRs \cite{AB2/SUU9CB_2019}, prior work has proposed crowdsourcing IDRs via DC insertion.  For example, to label the \textsc{reason} relation between the arguments "\textit{John missed the bus}" and "\textit{He was late to work.}", the DC "\textit{therefore}" could be inserted. In the initial proposal, crowd workers selected a DC from a fixed list, each corresponding to a unique IDR sense \cite{scholman-demberg-2017-crowdsourcing}.  While achieving high agreement with expert annotations, the method was tested on only $6$ IDR senses to avoid overwhelming workers with too many options. Choosing a DC is often context-dependent; for example, while "\textit{although}" and "\textit{even though}" are nearly interchangeable, "\textit{also}" versus "\textit{furthermore}" (both indicating \textsc{conjunction}) may depend on context. Workers might reject an appropriate sense if a DC feels contextually awkward.

To handle a broader range of IDRs, \citet{yung-etal-2019-crowdsourcing} proposed a two-step approach: first, workers freely type a DC that fits between two arguments; second, they select from a list of unambiguous DCs corresponding to their free-choice. For instance, if they type "\textit{while}" in the first step, they should choose between "\textit{at the same time"} and "\textit{in contrast}" in the second step, which are mapped to the relations \textsc{synchrony} and \textsc{contrast} respectively. This method was used to create the DiscoGeM 1.0 corpus, which contains $6,500$ English IDRs each annotated by $10$ workers \cite{scholman2022DiscoGeM}. Nonetheless, DiscoGeM 2.0, which extends the annotations to German, French, and Czech \cite{yung-etal-2024-discogem}, adopted the one-step forced-choice method: workers directly chose from $28$ DC choices, which were grouped by semantics and shuffled per worker to facilitate navigation and avoid positional bias.  The free- and forced- choice methods were reported to yield similar annotations, but the comparisons were based on a limited subset of items ($234$ in \citet{yung-etal-2019-crowdsourcing} and $18$ in \citet{yung-etal-2024-discogem}), with a restricted range of IDR senses.

Using a different crowd-annotation method, \citet{pyatkin-etal-2020-qadiscourse} crowdsourced discourse relations by instructing workers to create QA pairs from the provided text, e.g., "\textit{Q: What is the reason John was late? A: He missed the bus.}" Comparisons of QA-based and free-choice DC insertion methods show that both exhibit biases toward specific sense categories. In contrast to common attribution of method artifacts to degraded data quality \cite{gururangan-etal-2018-annotation,zhu-rzeszotarski-2024-get}, it was found that training on the complementary data collected by both methods enhanced the performance of IDR identification models \cite{pyatkin-etal-2023-design}.

\section{Annotation experiment}

We adopt the forced-choice approach to re-annotate the DiscoGeM 1.0 corpus, which was originally annotated using the free-choice approach. For this, an annotation interface was implemented based on the description of DiscoGeM 2.0 \cite{yung-etal-2024-discogem}. One representative DC was selected for each of the $28$ relations to be annotated. The selection was primarily based on the disambiguating DCs from the second step of the free-choice method,\footnote{the DC lexicon and per-worker annotations are available together with the corpus} 
 while ensuring they were sufficiently frequent and not highly context dependent. The complete list is shown in Table \ref{tab:dc_list} in the Appendix.

Following the procedure of DiscoGeM 1.0, native English-speaking crowd workers were recruited via the Prolific platform.  Based on the anonymous Prolific worker IDs, we invited the $199$ workers who contributed to DisocoGeM 1.0 to participate in the annotation task again. We assumed that they would not recall the texts they annotated three years ago and including them allows direct comparison of annotations from the same workers across both methods. Of these, $91$ workers took part again, and $73$ additional workers were recruited through a selection task. 

Out of the $6505$ items in DiscoGeM 1.0, $16$ duplicates were identified and removed. The remaining items were divided into batches of $20-25$, with each batch assigned to at least $10$ workers. The workers were awarded $\pounds 1.8 - \pounds 2.2$ per batch.

The quality of DiscoGeM 1.0 annotations was primarily controlled by a screening task that selected candidates achieving at least 50\% agreement with gold labels. During the data collection phase, annotation quality was monitored twice to identify and remove poorly performing annotators, while retaining their earlier annotations \cite{scholman2022DiscoGeM}. Similarly, we used an initial screening task to ensure annotation quality. However, to maximize the number of annotators participating in both tasks, we did not screen those who had contributed to DiscoGeM 1.0, nor did we conduct additional screening during the annotation process.

We compare the newly collected data against the original DiscoGeM 1.0.   In addition to analyzing label distributions from 10 workers per item, we compared aggregated annotations to highlight the differences.  The annotations were aggregated using the "Worker Agreement with Aggregate" (Wawa) algorithm, which weights each worker’s votes based on their overall agreement with the majority label \cite{ustalov2021general}.

\section{Results}

Table \ref{tab:agreement} presents the agreement between the annotations obtained by the two methods. We computed the averaged Jensen-Shanon divergense (JSD) between the label distributions of each item, as well as hard and soft agreement rates. Hard agreement measures matches between the single aggregated annotations, while soft agreement considers any overlap between annotations with over $20\%$ distribution a match \cite{pyatkin-etal-2023-design}. We also calculated the \textit{soft $\kappa$} scores, an inter-annotator agreement metric that accounts for the increased chance agreement in multi-label predictions \cite{marchal-etal-2022-establishing}.

\begin{table}[htpb]
    \centering
    \begin{tabular}{l|l|l}
         \hline
         \textit{inter-method comparison}& \multicolumn{2}{l}{free vs forced}\\
         \hline
         JSD (full dist.) & \multicolumn{2}{l}{$.527$}\\
         Hard agreement (single label) & \multicolumn{2}{l}{$.425$}\\
         Soft agreement (multi-labels) & \multicolumn{2}{l}{$.708$}\\
         Soft $\kappa$ (multi-labels)  & \multicolumn{2}{l}{$.663$}\\
         \hline
         \textit{intra-method comparison} & free &forced \\
         \hline
         Entropy & $0.353$&$0.460$\\
         Agreement (max. label dist) & $0.508$ &$0.404$ \\
         Per-item unique label count & $4.309$ & $6.275$ \\
         \hline
    \end{tabular}
    \caption{Annotation Agreement}
    \label{tab:agreement}
\end{table}

It can be observed that the inter-method agreement between single aggregated annotations is moderate, comparable to the accuracy of state-of-the-art IDR classification models \cite{costa2024multi,zeng-etal-2024-global}, but the agreement is substantially higher when multiple annotations are considered. This demonstrates that both methods are capable of annotating the same types of relations, which often co-occur with other relations.

The bottom half of Table \ref{tab:agreement} compares the agreement among the $10$ annotations per item in both methods. The forced-choice method shows higher averaged entropy in the per-item label distributions, indicating greater annotation uncertainty. In addition, the forced-choice approach yields smaller averaged per-item agreement (i.e., the proportion of the majority label) and a higher average number of unique annotations per item.  These results all indicate lower annotator agreement in the forced-choice approach. 

\begin{figure}[hptb]
\centering
  \includegraphics[width=\linewidth]{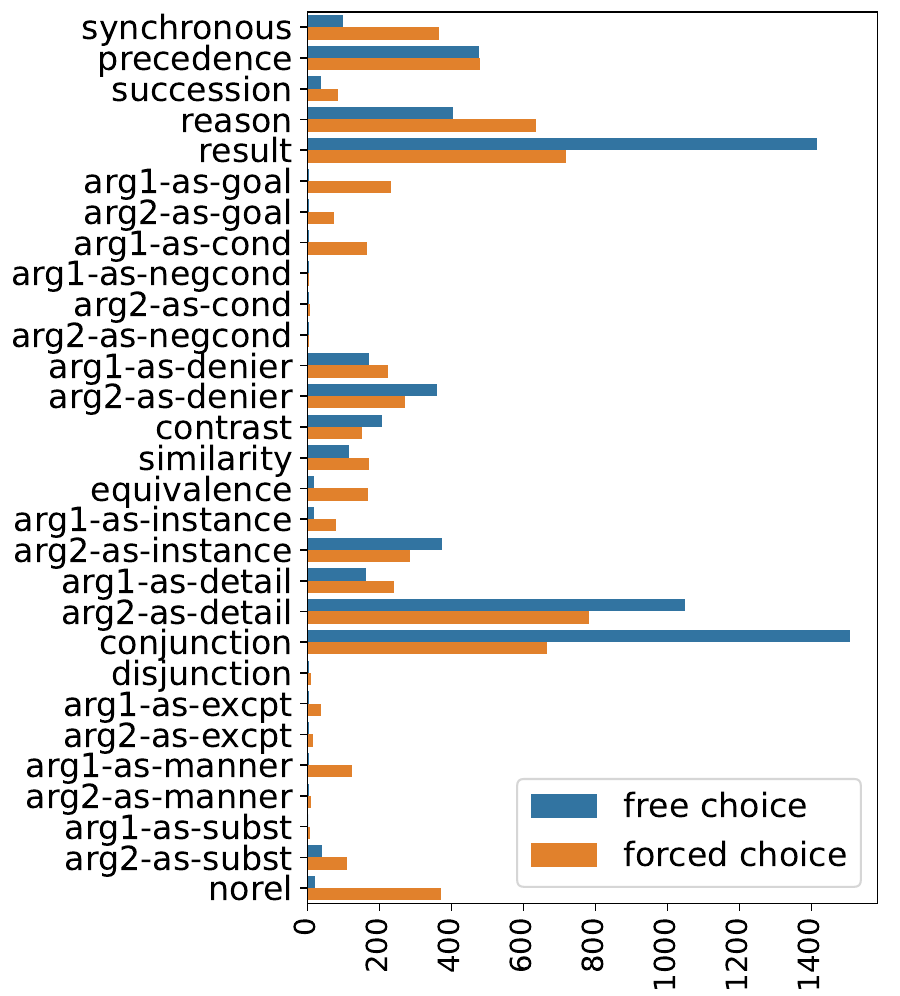}
  \caption {Distribution of the \textbf{unaggregated} annotations \label{fig:dist}}
\end{figure}
 Figure \ref{fig:dist} illustrates the overall distribution of the unaggregated annotations, computed by the sum of the normalized per-item distribution, since not all items have exactly $10$ annotations. The free-choice approach clearly converges on a narrower set of labels, while the forced-choice approach spans a wider range. Notably, \textsc{result} and \textsc{conjunction}, are selected twice as often in the free-choice method. 

\begin{figure}[h]
    
        \centering
        \includegraphics[width=\linewidth]{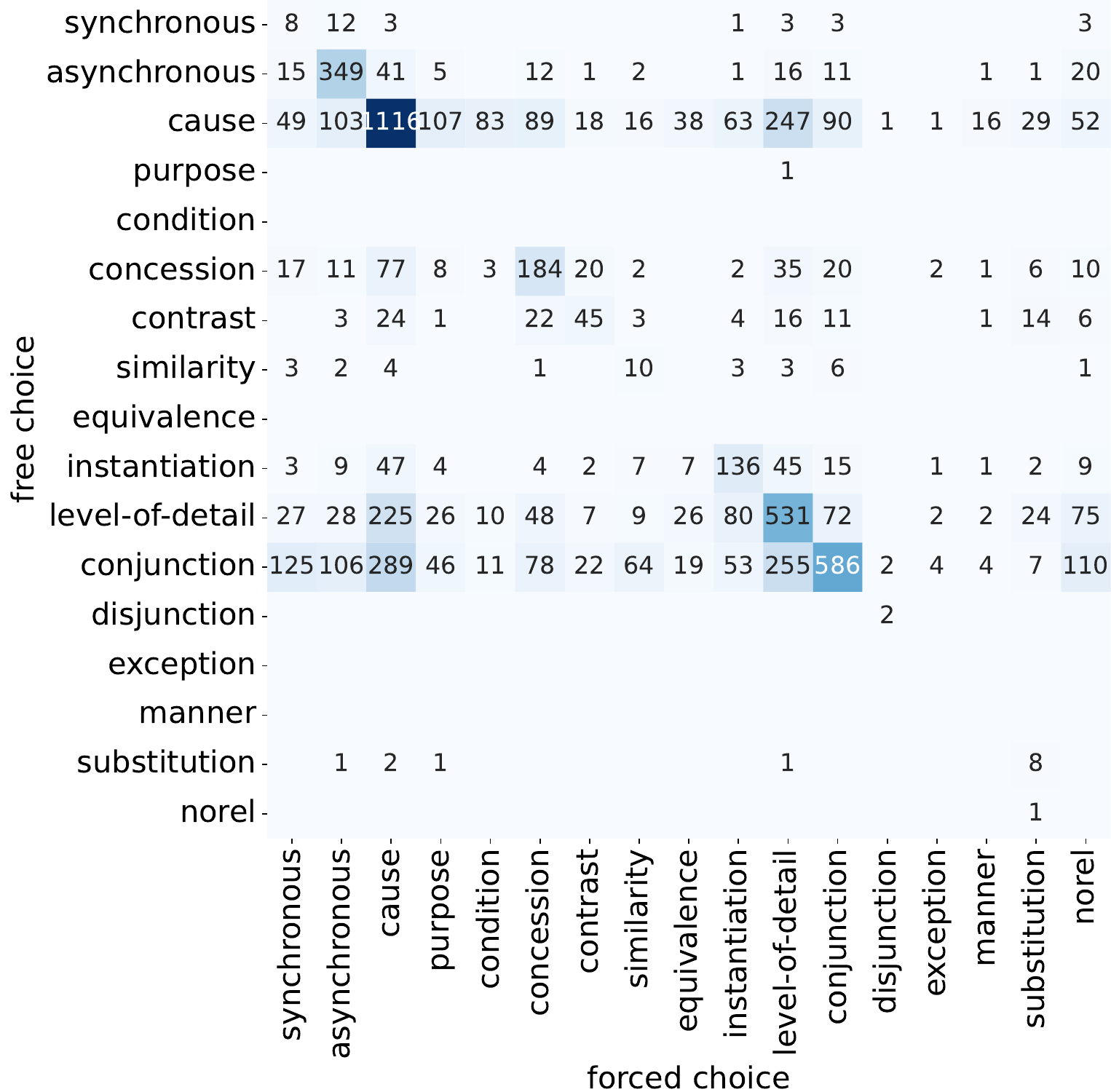} 
        \caption{Confusion matrix of the \textbf{aggregated} annotations from both methods, with labels merged at level-2 granularity}
        \label{fig:cm}
    \end{figure}

The trend is similar when focusing on the most agreed labels. Figure \ref{fig:cm} shows the alignment of the aggregated annotations from both methods. The annotations are grouped at level-2 granularity according to the PDTB sense hierarchy, e.g. \textsc{arg1-as-detail} and \textsc{arg2-as-detail} are grouped as \textsc{level-of-detail}.  Even though the darkest diagonal line in the confusion matrix indicates substantial agreement between annotations from both methods, many items labelled with \textsc{conjunction}, \textsc{causal}, and \textsc{arg2-as-detail} in the free-choice approach are now assigned to a range of other relations. While the aggregated annotations from the forced-choice approach cover all level-2 senses defined in the framework, half of these senses never appear in the aggregated annotations from the free-choice method.

Next, we directly compare the annotations of the same workers. In total, we identified $3,223$ annotations per method that were annotated by the same worker on the same item (spanning $2,542$ unique items and $91$ workers). The comparison of these annotations demonstrates a similar tendency as found in the re-annotation of the whole corpus, as shown in Figure \ref{fig:cm3223} in the Appendix - common relations like \textsc{conjunction} and \textsc{result} were annotated as other rarer relations in the forced-choice approach.

\begin{figure}[h]
        \centering
        \includegraphics[width=\linewidth]{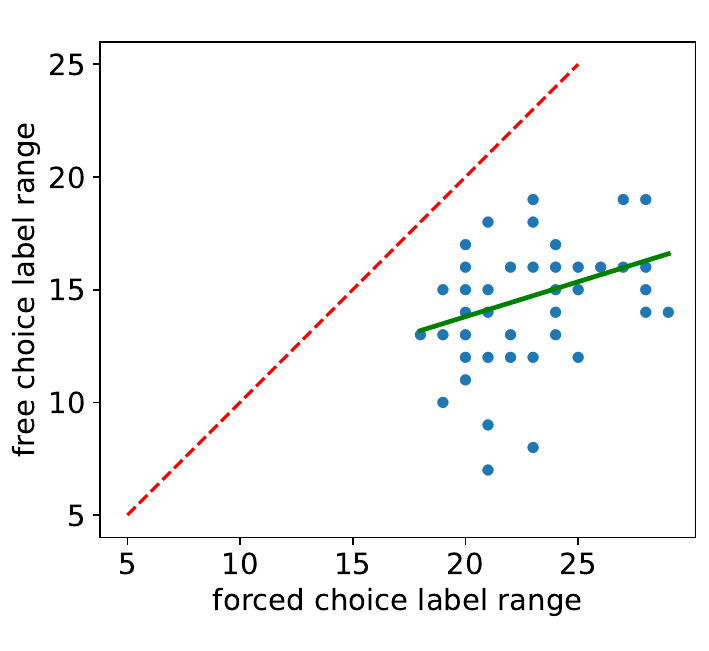} 
        \caption{Total number of \textbf{unique} relations annotated by the same workers on the same set of items  }\label{fig:range}
\end{figure}

Figure \ref{fig:range} plots the number of unique relations identified by workers who participated in both methods.
To ensure comparablility, results from workers who annotated fewer than $50$ items in either method or annotated items $3$ times more in one method than the other were excluded. This results in $60$ workers, who annotated on average $621$ and $525$ items in the free- and forced- choice methods respectively.

It shows that all workers identified a broader range of IDR senses using the forced-choice method, as indicated by all data points falling below the diagonal line. Furthermore, workers who could identify more sense types with the free-choice approach also identify more sense types in the forced-choice approach. This suggests individual differences in sensitivity to the subtle contrast in fine-grained discourse relations, with the forced-choice method further expanding the range of relations these workers could identify by presenting all possible options.

\section{Discussion and conclusion}

We examined the impact of two similar interfaces used to crowdsourcing IDR annotations. Using the free-choice approach, workers tend to select common IDR labels with higher inter-annotator agreement, while the forced-choice approach encouraged a larger variety of relations, including rare ones. Notably, both methods produce valid annotations, as evidenced by the high soft match agreement. Frequent senses can often be inferred alongside other senses, such as the \textsc{conjunction} sense in the examples in Figure \ref{fig:examples}. In these examples, the English forced-choice annotations align with other languages, despite being labeled as \textsc{conjunction} in the original free-choice annotations of DiscoGeM 1.0.

 \begin{figure} [htpb]
 \small
 1)
\framedtext{
Arg 1: It was because of this tiny piece of information that Ford Prefect was not now a whiff of hydrogen, ozone and carbon monoxide. He heard a slight groan.	

Arg2: By the light of the match he saw a heavy shape moving slightly on the floor. Quickly he shook the match out, reached in his pocket, found what he was looking for and took it out.
}
Aggregated annotation = \textsc{precedence}\\ 
(English, German, French, Czech forced-choice)\\
Aggregated annotation = \textsc{conjunction} \\
(English free-choice)\\\\

2) 
\framedtext{
Arg1: In yesterday's debate in the European Parliament some Members of this Parliament expressed worry that we were interfering in the internal affairs of a Member State. 

Arg2: Such a concern is misplaced.	The European Parliament has never been slow to comment on developments in Member States with which they disagree.
}
Aggregated annotation = \textsc{reason}\\ 
(English, German, French, Czech forced-choice)\\
Aggregated annotation = \textsc{conjunction} \\
(English free-choice)\\\\
 
3) 
 \framedtext{
Arg1: With a spring Gollum got up and started shambling off at a great pace. Bilbo hurried after him, still cautiously, though his chief fear now was of tripping on another snag and falling with a noise.	His head was in a whirl of hope and wonder. 

Arg2: It seemed that the ring he had was a magic ring: it made you invisible!
}
Aggregated annotation = \textsc{synchronous}\\ 
(English, German, French, Czech forced-choice)\\
Aggregated annotation = \textsc{conjunction} \\
(English free-choice)\\\\

\caption{Examples taken from DiscoGeM where the annotations by the forced- and free- choice approaches are alternative interpretations. The English forced-choice annotations come from the current study and those from the other languages come from DiscoGeM~2.0. The English free-choice annotations come from DiscoGeM~1.0.}
\label{fig:examples}
\end{figure}

High inter-annotator agreement is often linked to higher data quality. However, for inherently ambiguous tasks like IDR identification, we showed that higher-agreement annotations that converge on common labels are not always superior. 
Recognizing the method bias enables tailoring the approach to the annotation goal — whether to achieve consensus on a single label or capture diverse perspectives.  Since current IDR classification models often struggle with rare labels, datasets with more label variety may be more valuable. 
Still, distinguishing genuine perspectives from annotation errors is challenging. Minimal data cleaning, such as removing labels with very few votes, could be applied.

For corpus analysis, data should be collected consistently using the same method. Initial analysis reveals significant differences between the inter-annotator agreements of the English annotations in DiscoGeM 1.0 and the multilingual annotations in DiscoGeM 2.0, whereas the re-annotated data in this study aligns more closely with the other languages (e.g. averaged per-item agreement = $.508$/$.404$ (EN free-/forced-choice) $.410-.439$ (DE, FR, CS forced-choice), indicating the influence of the method bias. Our next step is to analyze the cross-lingual difference based on annotations collected with the same method.

\section*{Acknowledgements}
This project is supported by the German Research Foundation (DFG) under Grant SFB 1102 (``Information Density and Linguistic Encoding", Project-ID 232722074).




%

\bibliography{custom}
\appendix
\section{Appendix}
\label{sec:appendix}

\begin{table}[htpb]
    \centering \small
    \begin{tabular}{@{}l|l@{}}
    \hline
    level-2.level-3 IDR sense label & DC \\
    \hline\hline
    \textbf{Temporal} \\
\textsc{synchronous.synchronous} &at the same time \\
\textsc{asynchronous.precedence} &then \\
\textsc{asynchronous.succession} &after \\

    \hline
    \textbf{Contingency} \\
\textsc{cause.reason} &because \\
\textsc{cause.result} &as a result \\
\textsc{purpose.arg1-as-goal} &for that purpose \\
\textsc{purpose.arg2-as-goal} &so that \\
\textsc{condition.arg1-as-cond} &in that case \\
\textsc{condition.arg1-as-negcond} &if not \\
\textsc{condition.arg2-as-cond} &if \\
\textsc{condition.arg2-as-negcond} &unless \\

    \hline
    \textbf{Comparison} \\
\textsc{concession.arg1-as-denier} &even though \\
\textsc{concession.arg2-as-denier} &nonetheless \\
\textsc{contrast.contrast} &on the other hand \\
\textsc{comparison.similarity.similarity} &similarly \\

    \hline
    \textbf{Expansion}\\
\textsc{equivalence.equivalence} &in other words \\
\textsc{instantiation.arg1-as-instance} &this illustrates that  \\
\textsc{instantiation.arg2-as-instance} &for example \\
\textsc{level-of-detail.arg1-as-detail} &in short \\
\textsc{level-of-detail.arg2-as-detail} &in more detail \\
\textsc{conjunction.conjunction} &also \\
\textsc{disjunction.disjunction} &or \\
\textsc{exception.arg1-as-excpt} &other than that \\
\textsc{exception.arg2-as-excpt} &an exception is that \\
\textsc{manner.arg1-as-manner} &thereby \\
\textsc{manner.arg2-as-manner} &as if \\
\textsc{substitution.arg1-as-subst} &rather than \\
\textsc{substitution.arg2-as-subst} &instead \\

    \hline
\textsc{norel} & (no direct relation)    \\
    \hline

    \end{tabular}
    \caption{English DC choices used in the forced-choice DC insertion method}
    \label{tab:dc_list}
\end{table}

\begin{figure*}[htpb]
\centering
  \includegraphics[width=0.9\linewidth]{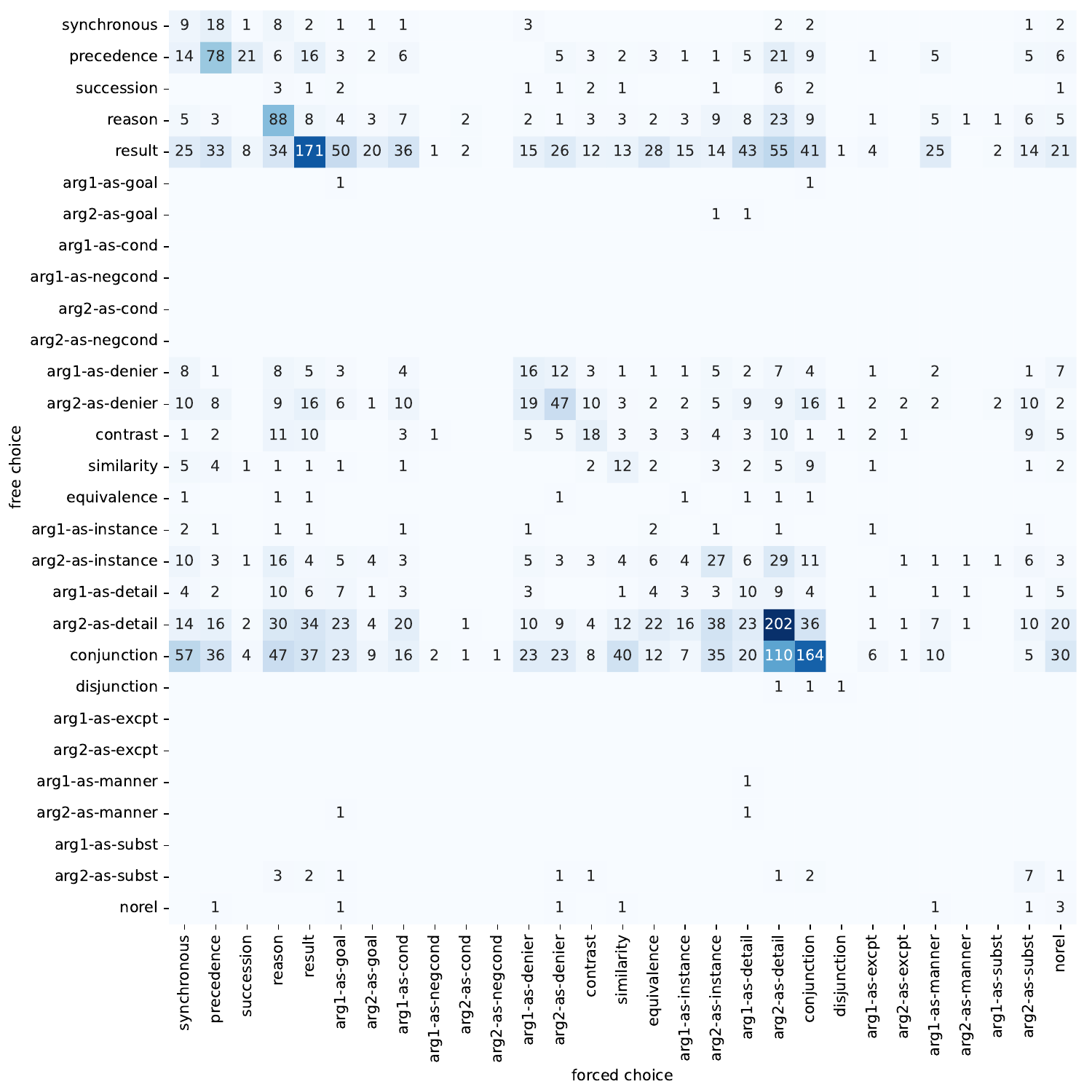}  
  \caption {Comparison between $3233$ annotations by the same workers on the same items using both methods\label{fig:cm3223} } 
\end{figure*}

\end{document}